\documentclass{article}

\usepackage{microtype}
\usepackage{graphicx}
\usepackage{fancyvrb}
\usepackage{subcaption}
\usepackage{booktabs} 
\usepackage{times}
\usepackage{graphicx}
\usepackage{amsmath}
\usepackage{amssymb}
\usepackage{booktabs}
\usepackage{multirow}
\usepackage[table]{xcolor}
\usepackage{hyperref}
\usepackage{fvextra}
\usepackage{tabularx}
\usepackage{microtype}
\usepackage{natbib}
\usepackage{array}
\usepackage{pifont}
\newcommand{\cmark}{\ding{51}}
\newcommand{\xmark}{\ding{55}}
\definecolor{better}{RGB}{212,237,218}
\definecolor{worse}{RGB}{248,215,218}
\definecolor{neutral}{RGB}{255,243,205}
\definecolor{hdrblue}{RGB}{46,64,87}
\definecolor{bad}{RGB}{248,215,218}
\usepackage{hyperref}

\usepackage[preprint]{icml2026}

\usepackage{amsmath}
\usepackage{amssymb}
\usepackage{mathtools}
\usepackage{amsthm}

\usepackage[capitalize,noabbrev]{cleveref}

\theoremstyle{plain}

\theoremstyle{definition}

\theoremstyle{remark}

\usepackage[textsize=tiny]{todonotes}

\begin{document}

\twocolumn[
  \icmltitle{Honest Lying: Understanding Memory Confabulation in  Reflexive Agents}



\icmlsetsymbol{equal}{*}

\begin{icmlauthorlist}
  \icmlauthor{Prakhar Dixit}{umbc}
  \icmlauthor{Sadia Kamal}{umbc}
  \icmlauthor{Tim Oates}{umbc}
\end{icmlauthorlist}

\icmlaffiliation{umbc}{Department of Computer Science, University of Maryland Baltimore County, Baltimore, MD, USA}

\icmlcorrespondingauthor{Prakhar Dixit}{pdixit1@umbc.edu}
\icmlcorrespondingauthor{Sadia Kamal}{sadia1402@umbc.edu}
\icmlcorrespondingauthor{Tim Oates}{oates@cs.umbc.edu}

\icmlkeywords{Machine Learning, Multi-Agent Systems, Reflexive Agents}

\vskip 0.3in
]

\printAffiliationsAndNotice{}

\begin{abstract}
Reflexion-style agents rely on self-generated reflections as memory, implicitly assuming that agents can accurately diagnose their own failures. We show that this assumption can fail systematically: across ALFWorld and HumanEval, agents store confident but incorrect interpretations of the task and continue acting on them across trials, even though the environment resets to the correct task each time. 
We call this failure mode \emph{memory confabulation} and introduce the \emph{Reflection Repetition Rate} (RRR), a log-based metric that detects repeated reliance on incorrect reflective content. 
Using RRR, we identify 16 frozen environments in ALFWorld, where 0 of 121 reflections mention the correct target object, and 4 analogous cases in HumanEval. Our mitigation replaces open-ended self-diagnosis with programmatic extraction of trajectory-level failure signals, increasing correct object mention from 0\% to 86\%, reducing RRR from 0.64 to 0.10, and solving 3 of 16 frozen ALFWorld environments, suggesting that reflective memory can reinforce false beliefs rather than correct them.
\end{abstract}

\section{Introduction}

Foundation-model agents increasingly operate in settings where they must learn from experience across repeated trials~\cite{durante2025interactive}. Reflexion~\citep{shinn2023reflexion} is a representative approach: after failure, an agent writes a natural-language reflection and retrieves it in later attempts. This design assumes that self-generated reflection produces useful failure diagnoses. When the assumption holds, the agent can convert a failed action into a reusable lesson.

We show that this assumption can fail systematically. Reflexion agents may write confident but incorrect accounts of the task, store them as memory, and reuse them across trials even when the environment re-presents the correct task at every reset. We call this failure mode \textbf{memory confabulation}: persistent, self-reinforcing false beliefs written into reflective memory and acted upon despite contradictory task evidence. The term follows cognitive accounts of confabulation as a failure of reality monitoring, where internally generated information is mistaken for observed information~\citep{johnson1998false,schnider2001spontaneous,chrobak2009cognitive}.

Memory confabulation differs from hallucination. Hallucination is typically a single-generation error, while memory confabulation is a multi-trial failure: the false content is stored, retrieved, acted upon, and reinforced by later reflections~\cite{ji2023survey,maynez2020faithfulness}. This distinction matters for agent design because the error is not merely produced once; it becomes part of the agent's future context. Therefore, we ask: \emph{can reflective agents rely on their own self-generated lessons, or does verbal self-diagnosis create false beliefs in memory that persist across trials and degrade rather than improve performance?}. \textbf{Our main contributions are}:
\begin{itemize}
    \item We identify and operationalize \textbf{memory confabulation}: false self-diagnoses written into reflective memory and reused across trials despite contradictory task evidence.

    \item We introduce the \textbf{Reflection Repetition Rate (RRR)}, a log-based diagnostic for frozen reflective memory that strongly correlates with trials-to-solve in ALFWorld~\cite{shridhar2020alfworld} ($r = 0.808$).

    \item We provide cross-domain evidence from ALFWorld and HumanEval~\cite{chen2021evaluating}, finding that 0/121 reflections in 16 frozen ALFWorld environments mention the correct target object and that 4 HumanEval problems repeat near-identical wrong diagnoses.

    \item We show through a no-memory ablation that reflective memory can be actively harmful in some environments the agent can otherwise solve.

    \item We propose \textbf{programmatic feedback extraction}, which replaces open-ended self-diagnosis with parsed trajectory-level failure signals, raising correct object mention from 0\% to 86\%, reducing RRR from 0.64 to 0.10, and solving 3/16 frozen ALFWorld environments.
\end{itemize}
\section{Background}

\paragraph{Reflexion.}
Reflexion~\citep{shinn2023reflexion} extends ReAct-style 
agents~\citep{yao2022react} with a verbal reinforcement mechanism. After each failed 
task attempt, a language model generates a natural-language self-critique which is 
prepended to the agent's context on subsequent attempts. No gradient updates are 
performed; learning is entirely mediated by the context. Reflexion achieves 91\% 
pass@1 on HumanEval versus 80\% for GPT-4 \cite{hurst2024gpt} without reflection, demonstrating the 
power of the mechanism when it works correctly. 
However, Reflexion assumes the reflection step produces causally correct diagnoses. 
We show this assumption fails systematically when the feedback signal is binary 
(pass/fail) and the task requires multi-step manipulation.
 
\paragraph{Memory in LLM agents.}
Recent surveys~\citep{zhang2025survey,du2026memoryautonomousllmagentsmechanisms} formalize agent memory as a 
write--manage--read loop and identify reflective self-improvement as a distinct 
memory mechanism family. \citet{du2026memoryautonomousllmagentsmechanisms} note that the central risk of 
reflective memory is self-reinforcing error: if the agent falsely concludes that an 
approach always fails, it will never test that approach again. Our work 
operationalizes this risk empirically and identifies the specific triggering 
condition: binary feedback that prevents causal diagnosis \cite{zhang2026feedback}.
 
\paragraph{ExpeL and rule-library agents.}
ExpeL~\citep{zhao2024expel} extends Reflexion with a shared rule library: 
experience is distilled into globally-applicable rules via unconstrained LLM 
critique on failure trajectories. This shares the same structural vulnerability as 
Reflexion, but with amplified consequences: where Reflexion confabulates per-task 
reflections, a confabulated rule with two AGREE votes becomes entrenched and is 
applied across every evaluation environment. We discuss this generalization in 
Section~\ref{sec:discussion}.
 
\paragraph{Hallucination and confabulation.}
Hallucination in LLMs has been extensively studied as a single-generation failure 
\citep{Ji_2023,maynez2020faithfulnessfactualityabstractivesummarization}. Memory confabulation is structurally 
distinct: the false content is stored, retrieved, and acted upon persistently across 
multiple trials.
 
\section{Problem Formulation}
 
\subsection{Operational Definition}
 
Let an agent operate on task $\tau$ over trials $t = 0, 1, \ldots, T$. After each 
failure at trial $t$, the agent generates reflection $r_t$ from the trajectory and 
stores it: $M_{t+1} = M_t \cup \{r_t\}$. At trial $t+1$, the agent retrieves from 
$M_{t+1}$ to inform its actions.
 
\paragraph{Definition (Memory Confabulation).}
A reflection $r_t$ is \emph{confabulated} if it fails to mention the correct target 
object of task $\tau$, i.e., $\mathrm{obj}(\tau) \notin r_t$, where $\mathrm{obj}(\tau)$ 
extracts the target object from the task description presented to the agent at the 
start of every episode.

This definition is operationalizable from existing logs without new experiments, 
using only the gamefile directory name (which encodes task structure) and the stored 
reflection text.

\subsection{Reflection Repetition Rate (RRR)}
 
For an environment with memory $M = \{r_0, \ldots, r_n\}$, we define:
\begin{equation}
\mathrm{RRR} = \frac{\left|\{r_i : i \geq 1,\ \exists j < i,\ \mathrm{sim}(r_i, r_j) \geq 0.85\}\right|}{|M| - 1}
\end{equation}
where $\mathrm{sim}$ is SequenceMatcher string similarity \cite{musser2008fastgenericsequencematching}. $\mathrm{RRR} = 0$ 
indicates all reflections are novel; $\mathrm{RRR} = 1$ indicates all reflections 
are near-copies of earlier ones. We use 0.85 as the similarity threshold for 
near-duplication.

We define an environment as exhibiting \textbf{frozen reflective memory} when $\mathrm{RRR} \geq 0.5$, meaning that at least half of the reflections after the
first are near-duplicates of earlier reflections. This threshold identifies cases
where reflective memory stops evolving across trials and repeatedly reuses the
same content. We use the term \emph{frozen environment} as shorthand for an
environment whose reflective memory is frozen, not to imply that the underlying
task or simulator state is fixed.
 
\subsection{Two Failure Categories}
 
Our analysis distinguishes two categories among frozen environments ($\mathrm{RRR} \geq 0.5$):
 
\textbf{Memory-harmful}: removing memory allows the agent to solve the task faster, 
proving stored reflections were actively misleading.
 
\textbf{Task-hard}: the agent fails even without memory, indicating a capability 
gap independent of memory quality.
 
\section{Evidence of Memory Confabulation}
\label{sec:evidence}

\subsection{RRR Analysis}

Using pre-existing Reflexion run logs (134 environments, 15 trials, gpt-3.5-turbo \cite{brown2020language}
on ALFWorld \texttt{eval\_out\_of\_distribution}), we compute RRR for all 50
environments that required at least one reflection. We find 16 of 50 environments
(32\%) exhibit frozen memory (RRR $\geq 0.5$). Frozen environments required an
average of 7.6 trials to solve, versus 1.5 trials for environments with diverse,
evolving reflections. The Spearman correlation between RRR and trials-to-solve is
$r = 0.808$ ($p < 0.0001$), suggesting that frozen memory is not incidental but
is a reliable predictor of agent failure.

\subsection{Task-Object Confabulation}

We extracted the target object from each frozen environment's gamefile directory
name and checked whether each reflection mentioned that object. For example,
\begin{Verbatim}[
  fontsize=\footnotesize,
  breaklines=true,
  breakanywhere=true,
  xleftmargin=1em
]
pick_cool_then_place_in_recep-Mug-None-CoffeeMachine-10
\end{Verbatim}
encodes object \texttt{Mug} and destination \texttt{CoffeeMachine}.
Across all 16 frozen environments, 0 of 121 reflections mention the correct
target object.

\begin{figure}[t]
    \centering
    \includegraphics[width=\linewidth]{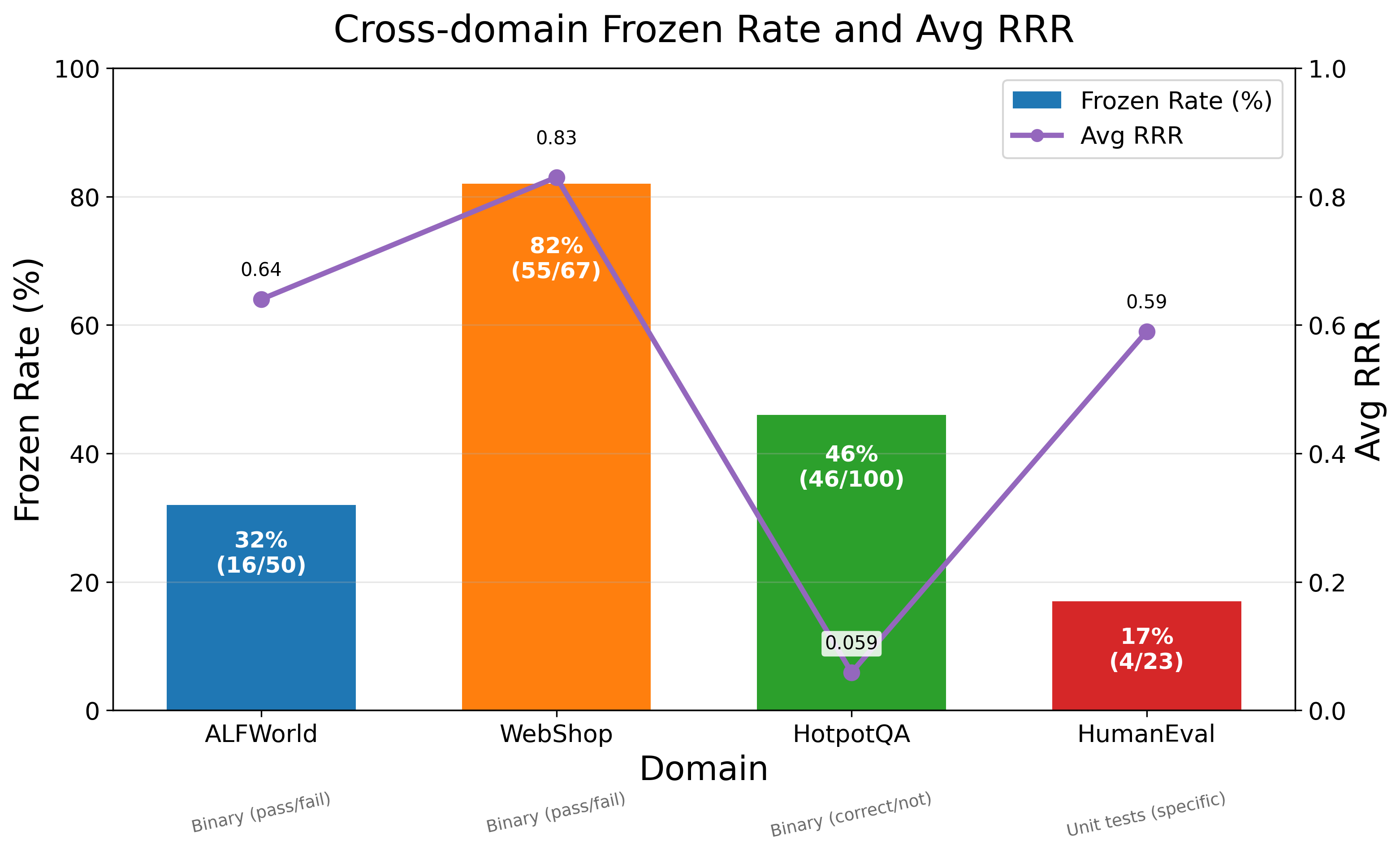}
    \caption{Cross-domain frozen rate and average RRR by feedback type. Binary outcome-level feedback is associated with higher frozen-memory rates, while more specific feedback supports targeted self-correction.}
    \label{fig:crossdomain-feedback}
\end{figure}

The most striking case is \texttt{env\_22} (task: put a cool Mug in
CoffeeMachine), where all 14 reflections reference \textit{tomato} and
\textit{microwave}---a completely different task. The agent fabricated a false
task identity after trial 0 and pursued it for 14 consecutive trials, ignoring
the correct task description re-presented at every episode reset.

A second case, \texttt{env\_35} (task: examine the Mug with the desklamp),
illustrates two compounding confabulation patterns: \emph{location
confabulation}, where the agent follows a search sequence from a previous
environment layout across trials 0--2, and \emph{action confabulation}, where
the agent attempts to use an object without navigating to it because its memory
incorrectly anchors its believed location.

Two distinct confabulation patterns emerge across the 16 frozen environments:
\begin{itemize}
    \item \textbf{Full task substitution}: both object and destination are
    replaced (\texttt{env\_22}, \texttt{env\_20}, \texttt{env\_41}).
    \item \textbf{Object substitution only}: the destination is correctly
    remembered but the object is replaced (\texttt{env\_118},
    \texttt{env\_113}, \texttt{env\_106}).
\end{itemize}

\subsection{Cross-Domain Replication}

We replicate the frozen memory analysis across three additional Reflexion
domains available in the same repository, requiring no new experiments or API
calls. Results are shown in Figure~\ref{fig:crossdomain-feedback} and
Table~\ref{tab:crossdomain}.

\begin{table}[t]
\centering
\caption{Cross-domain frozen memory rates by feedback type. All results are computed from pre-existing Reflexion run logs.}
\label{tab:crossdomain}
\small
\setlength{\tabcolsep}{3pt}
\renewcommand{\arraystretch}{1.15}
\begin{tabular}{l l c c}
\toprule
\textbf{Domain} & \textbf{Feedback} & \textbf{Frozen} & \textbf{Avg RRR} \\
\midrule
ALFWorld  & Binary        & 32\% (16/50)  & 0.64  \\
WebShop   & Binary        & 82\% (55/67)  & 0.83  \\
HotpotQA  & Binary        & 46\% (46/100) & 0.059 \\
HumanEval & Unit tests    & 17\% (4/23)   & 0.59  \\
\bottomrule
\end{tabular}
\end{table}
The results reveal a consistent pattern: domains with binary, outcome-level
feedback (ALFWorld, WebShop \cite{yao2022react}, HotpotQA \cite{yang2018hotpotqa})  exhibit substantially higher frozen
rates and lower RRR than domains with specific, step-level feedback (HumanEval).
HotpotQA is the most acute case. Despite seven trials of reflection on 100
multi-hop questions, the agent corrected a previously wrong answer only 5.9\%
of the time per trial transition---compared to 64\% for ALFWorld and 83\% for
WebShop. This is because binary correct/wrong feedback on open-ended reasoning
provides no signal about \emph{which step} of a multi-hop chain failed, leaving
the agent unable to target its self-correction. The 46\% frozen rate (46 of 100
questions never answered correctly across seven trials) reflects this
ceiling effect.

WebShop confabulation manifests differently from ALFWorld. Rather than
substituting the wrong task object, agents exhibit \emph{symptom
confabulation}: 56\% (121/218) of frozen reflections describe what went wrong
(``I clicked the wrong item'') without diagnosing why---which size, color, or
price constraint was violated. This is a distinct surface form of the same root
cause: binary feedback contains no step-level information, so reflections
recapitulate failure without identifying it.

HumanEval's comparatively low persistent failure rate (17\%) supports the contrastive hypothesis. Unit test feedback names the exact
assertion that failed, giving the agent a precise error to reason about.
Reflections in this domain are correspondingly targeted, and the agent
self-corrects at nearly ten times the rate observed on HotpotQA.
 
\subsection{Evidence of Memory Confabulation}
 
Correlation between RRR and trials-to-solve is not causal evidence; frozen environments may simply be harder tasks. To establish causality, we re-ran all 16 
frozen environments with memory completely wiped before each trial (gpt-3.5-turbo,max 10 trials, ALFWorld \texttt{eval\_out\_of\_distribution}).
 
\paragraph{Results.}
The 16 frozen environments split cleanly into two categories:
 
\textbf{Memory-harmful (2/16):} \texttt{env\_31} (\texttt{look\_at\_obj\_in\_light}) 
and \texttt{env\_97} (\texttt{look\_at\_obj\_in\_light}). Without memory, both solve 
in 1 trial. With standard Reflexion memory: 7 and 8 trials respectively. 
\textbf{Delta: $+6$ and $+7$ trials.}
 
\textbf{Task-hard (14/16):} All \texttt{pick\_heat}, \texttt{pick\_cool}, 
\texttt{pick\_clean}, and \texttt{pick\_and\_place} environments. The agent fails 
within 10 trials even without memory, indicating a capability gap independent of 
memory quality.
 
This finding establishes that memory confabulation has two distinct effects: it 
directly causes failure in tasks the agent could otherwise solve (memory-harmful 
category), and it compounds an existing capability gap in hard tasks (task-hard 
category).
 
\section{Mitigation and Results}
 
We test two mitigation strategies on all 16 frozen environments. Both target the 
same root cause: binary feedback preventing accurate causal diagnosis.
 
\subsection{Grounded Reflection}
 
We modified the reflection prompt to require a structured three-part response:
\begin{Verbatim}[
fontsize=\footnotesize,
breaklines=true,
breakanywhere=true,
frame=single,
framesep=2mm
]
FAILED STEP:
  [exact action + environment response]
ROOT CAUSE:
  [one sentence why that action failed]
NEW PLAN:
  [specific plan naming exact objects]
\end{Verbatim}
This forces the agent to quote a specific trace step before planning. Result: 
grounded reflection matches no-memory performance on look\_at\_obj environments 
(TTS = 1), confirming confabulation is prevented, but produces no improvement on 
task-hard environments.
 
\subsection{Programmatic Feedback Extraction}
 
Motivated by the cross-domain finding that unit-test feedback produces 17\% frozen 
rate versus 32--82\% for binary feedback, we implemented a trajectory parser that 
extracts failure steps programmatically rather than asking the agent to self-diagnose. 
The parser identifies: (1) actions that received ``Nothing happens'' responses, and 
(2) repeated identical actions indicating a loop. These are injected directly into 
the reflection prompt. The full prompt template can be found in Appendix~\ref{sec:prompt}. 
\begin{Verbatim}[
fontsize=\footnotesize,
breaklines=true,
breakanywhere=true,
frame=single,
framesep=2mm
]
TASK: [task description]
FAILURES FROM TRACE:
  1. Action: put mug 1 in coffeemachine 1
     Response: Nothing happens.
\end{Verbatim}
This replicates what unit tests do for HumanEval: providing grounded failure 
evidence instead of requiring self-diagnosis.

\begin{figure}
    \centering
    \includegraphics[width=1\linewidth]{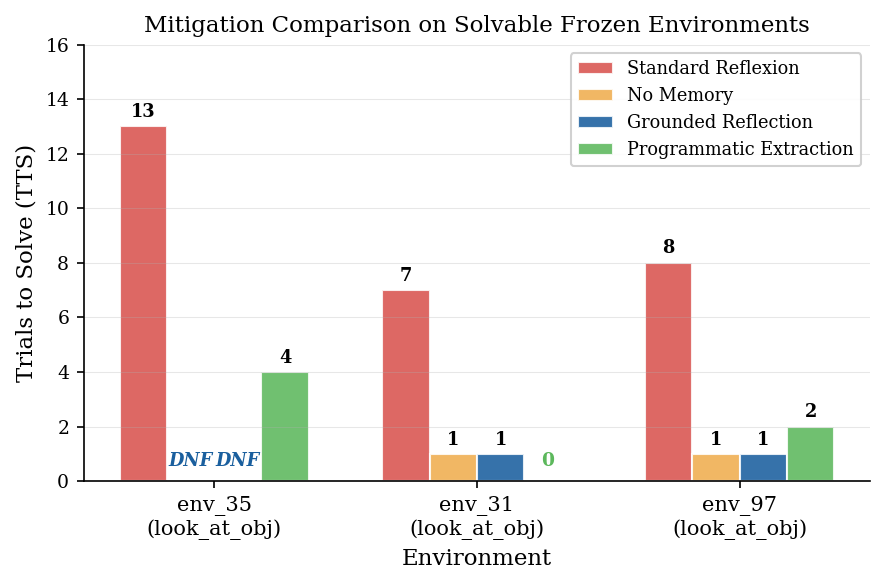}
    \caption{Mitigation comparison across frozen ALFWorld environments. Programmatic feedback extraction substantially reduces repeated reliance on incorrect reflections, while solving additional environments beyond no-memory and grounded-reflection baselines.}
    \label{fig:mitigation-comparison}
\end{figure}

\subsection{ALFWorld Results}
Figure~\ref{fig:mitigation-comparison} summarizes the mitigation results across
all five conditions. Table~\ref{tab:mitigation} presents the full comparison across five conditions 
on all 16 frozen ALFWorld environments. Under standard Reflexion, all 16 
environments exhibit fully confabulated memory: 0 of 121 reflections mention 
the correct target object, and none solve within the 10-trial budget used 
for all mitigation conditions. Programmatic feedback extraction raises the 
correct object mention rate to 134/156 (86\%) and reduces average RRR from 
0.64 to 0.10, confirming that the mitigation breaks frozen memory patterns 
even on environments that remain unsolvable due to capability gaps. The 22 
remaining mention misses are concentrated in \texttt{env\_4}, where the 
search for \texttt{SoapBar} produces partial regex matches.

Removing memory entirely solves 2/16 environments (\texttt{env\_31}, 
\texttt{env\_97}), both \texttt{look\_at\_obj} tasks that complete in 1 trial 
without confabulated reflections. Grounded reflection matches this baseline. 
Programmatic extraction solves 3/16, uniquely resolving \texttt{env\_35} 
which all other conditions fail on. Replication with gpt-4o-mini eliminates 
task-identity confabulation (100\% object mention rate) but solves only 2/16, 
identical to the no-memory baseline, confirming that confabulation prevention 
and task-level capability are independent failure axes.
 
\begin{table*}[t]
\centering
\caption{Trials-to-solve across five conditions for all 16 frozen environments. 
All mitigation conditions use a 10-trial budget. DNF means the agent did not 
finish within 10 trials. Orig column shows baseline TTS from the original 
15-trial Reflexion logs where all 16 environments exhibited confabulated memory 
(0/121 reflections mention the correct object).}
\label{tab:mitigation}
\renewcommand{\arraystretch}{1.15}
\setlength{\tabcolsep}{4pt}
\small
\resizebox{\textwidth}{!}{%
\begin{tabular}{llccccc}
\toprule
\textbf{Env} & \textbf{Task Type} & \textbf{Orig(All Confabulated)} & \textbf{No Mem} & 
\textbf{Grounded} & \textbf{Enriched} & \textbf{gpt-4o-mini} \\
\midrule
env\_35  & look\_at\_obj    & \cellcolor{bad}13$^\dagger$ & DNF & DNF & \cellcolor{better}\textbf{4} & DNF \\
env\_31  & look\_at\_obj    & \cellcolor{bad}7$^\dagger$  & 1   & 1   & \cellcolor{better}\textbf{1} & 1 \\
env\_97  & look\_at\_obj    & \cellcolor{bad}8$^\dagger$  & 1   & 1   & \cellcolor{better}\textbf{2} & 1 \\
\midrule
env\_4   & pick\_clean      & \cellcolor{bad}3$^\dagger$  & DNF & DNF & \cellcolor{neutral}DNF & DNF \\
env\_20  & pick\_and\_place & \cellcolor{bad}6$^\dagger$  & DNF & DNF & \cellcolor{neutral}DNF & DNF \\
env\_22  & pick\_cool       & \cellcolor{bad}14$^\dagger$ & DNF & DNF & \cellcolor{neutral}DNF & DNF \\
env\_40  & pick\_and\_place & \cellcolor{bad}6$^\dagger$  & DNF & DNF & \cellcolor{neutral}DNF & DNF \\
env\_41  & pick\_heat       & \cellcolor{bad}13$^\dagger$ & DNF & DNF & \cellcolor{neutral}DNF & DNF \\
env\_56  & pick\_cool       & \cellcolor{bad}2$^\dagger$  & DNF & DNF & \cellcolor{neutral}DNF & DNF \\
env\_77  & pick\_heat       & \cellcolor{bad}6$^\dagger$  & DNF & DNF & \cellcolor{neutral}DNF & DNF \\
env\_80  & pick\_heat       & \cellcolor{bad}5$^\dagger$  & DNF & DNF & \cellcolor{neutral}DNF & DNF \\
env\_82  & pick\_heat       & \cellcolor{bad}7$^\dagger$  & DNF & DNF & \cellcolor{neutral}DNF & DNF \\
env\_86  & pick\_two\_obj   & \cellcolor{bad}4$^\dagger$  & DNF & DNF & \cellcolor{neutral}DNF & DNF \\
env\_106 & pick\_and\_place & \cellcolor{bad}6$^\dagger$  & DNF & DNF & \cellcolor{neutral}DNF & DNF \\
env\_113 & pick\_cool       & \cellcolor{bad}9$^\dagger$  & DNF & DNF & \cellcolor{neutral}DNF & DNF \\
env\_118 & pick\_and\_place & \cellcolor{bad}12$^\dagger$ & DNF & DNF & \cellcolor{neutral}DNF & DNF \\
\bottomrule

\textit{Solved}              & & 0/16 & 2/16 & 2/16 & \textbf{3/16} & 2/16 \\
\textit{Object mention rate} & & 0\%  & ---  & ---  & \textbf{86\%} & \textbf{100\%} \\
\textit{Avg RRR}             & & 0.64 & ---  & ---  & \textbf{0.10} & 0.53 \\
\bottomrule
\end{tabular}%
}
\end{table*}
\paragraph{env\_35 case study.}
This environment is the most informative result. It was DNF under both no-memory 
and grounded reflection baselines, but programmatic feedback extraction solved it 
in 4 trials. The trajectory shows that trials 0--2 were dominated by a frozen 
location plan derived from a previous task layout (location confabulation). 
Programmatic extraction broke this freeze by surfacing the specific ``Nothing 
happens'' response, enabling the agent to search for an alternative desklamp and 
solve the task in 33 steps at trial 4.
 
\subsection{HumanEval Results}
\label{sec:humaneval}
 
To assess whether programmatic feedback extraction generalizes across task types, we 
apply the same principle to HumanEval code generation, a structurally different 
domain where the agent writes Python functions and receives unit-test feedback. 
The extraction is domain-adapted: instead of parsing \texttt{Nothing happens}, we 
parse the failing \texttt{assert} statement and error type from the test output.
 
\paragraph{Confabulation in code generation.}
We identify 4 frozen HumanEval problems ($\mathrm{RRR} \geq 0.5$) from the 
pre-existing Reflexion logs. In all four cases, the agent produces near-identical 
reflections across 9 trials without diagnosing the specific failing input. For 
example, HumanEval/32 (\texttt{find\_zero} binary search) generates the same 
diagnosis --- ``does not update lower\_bound and upper\_bound'' --- across all 9 
reflections, never identifying \emph{which} input triggers the infinite loop 
or \emph{why} the bounds fail to converge.
 
\paragraph{Domain-adapted extraction.}
For HumanEval, the programmatic extractor identifies:
\begin{enumerate}
    \item The failing \texttt{assert} statement (e.g., 
    \texttt{assert candidate(1000) == "1"})
    \item The error type (\texttt{AssertionError}, 
    \texttt{TypeError}, etc.) and its associated traceback message
\end{enumerate}
These are injected into the reflection prompt in place of the agent's self-diagnosis, 
mirroring the ALFWorld \texttt{Nothing happens} extraction. The agent is 
\emph{told} which specific test case failed rather than asked to infer it.
 
Table~\ref{tab:humaneval} shows results across the 4 frozen problems. 
 
\begin{table}[t]
\centering
\caption{HumanEval results with programmatic feedback extraction on 4 frozen problems.}
\label{tab:humaneval}
\renewcommand{\arraystretch}{1.1}
\setlength{\tabcolsep}{3pt}
\footnotesize
\resizebox{\columnwidth}{!}{%
\begin{tabular}{lccccc}
\toprule
\textbf{Task} & \textbf{Orig} & \textbf{Enriched} & \textbf{Orig} & 
\textbf{New} & \textbf{Error} \\
 & \textbf{Solved} & \textbf{Solved} & \textbf{RRR} & \textbf{RRR} & 
\textbf{Mention} \\
\midrule
HumanEval/32 & \xmark & \xmark & 0.50 & 0.75 & 9/9 \\
HumanEval/77 & \cmark & \xmark & 0.50 & 1.00 & 9/9 \\
HumanEval/84 & \xmark & \cmark & 0.75 & 0.00 & 0/0 \\
HumanEval/87 & \cmark & \cmark & 0.62 & 0.00 & 0/0 \\
\midrule
\textit{Summary} & 2/4 & 2/4 & 0.59 & \textbf{0.44} & \textbf{18/18} \\
\bottomrule
\end{tabular}%
}
\end{table}
 
The key findings on HumanEval mirror the ALFWorld results:
 
\textbf{Grounding works}: 18/18 (100\%) of generated reflections mention the 
specific error type, compared to the near-identical vague diagnoses produced by 
standard Reflexion. This mirrors the 0/121 $\to$ 134/156 object mention improvement 
on ALFWorld.
 
\textbf{RRR reduces}: Average RRR drops from 0.59 to 0.44, confirming that 
programmatic extraction breaks frozen memory patterns in code generation.
 
\textbf{Capability gaps persist}: HumanEval/32 and /84 fail even with grounded 
reflections --- the binary search implementation error and digit-sum binary 
conversion require algorithmic insight that no memory-level intervention provides. 
This mirrors the DNF pattern on ALFWorld pick\_* tasks.
 
\textbf{One regression}: HumanEval/77 regressed from solved to unsolved. The 
structured reflection prompt disrupted a working solution strategy --- a cost 
consistent with the memory-harmful vs.\ task-hard finding on ALFWorld, and a 
reminder that memory interventions carry risk even when they improve grounding.
 
Note that HumanEval/84 and /87 show 0/0 grounded reflections because both solved 
at trial 0 (no reflections generated). Their RRR drops to 0.00 as a result, not 
due to the mitigation.
 
\paragraph{Cross-domain parallel.}
Table~\ref{tab:comparison} summarises the structural parallel between the two 
mitigation experiments.
 
\begin{table}[t]
\centering
\caption{Structural parallel between ALFWorld and HumanEval mitigation experiments.}
\label{tab:comparison}
\renewcommand{\arraystretch}{1.15}
\setlength{\tabcolsep}{3pt}
\footnotesize
\begin{tabularx}{\columnwidth}{lXX}
\toprule
 & \textbf{ALFWorld} & \textbf{HumanEval} \\
\midrule
Task type        & Household navigation          & Code generation \\
Feedback type    & Binary (pass/fail)            & Unit tests (specific) \\
Failure signal   & \texttt{Nothing happens}      & \texttt{AssertionError} \\
Extracted info   & Failed action and response    & Failing assert and error type \\
Confab.\ pattern & Wrong task object             & Vague or wrong diagnosis \\
Frozen rate      & 32\% (16/50)                  & 17\% (4/23) \\
Mention rate     & 0\% $\rightarrow$ 86\%        & 0\% $\rightarrow$ 100\% \\
RRR change       & 0.64 $\rightarrow$ \textbf{0.10} & 0.59 $\rightarrow$ 0.44 \\
Solved           & 3/16                          & 2/4 \\
Capability gaps  & 13/16 DNF                     & 2/4 DNF \\
\bottomrule
\end{tabularx}
\end{table}

\section{Discussion}
\label{sec:discussion}
 
\subsection{The Feedback Granularity Hypothesis}
 
Our cross-domain analysis points to a clear mechanism: binary feedback (pass/fail) 
gives the agent no information about which step in a multi-step trajectory was wrong. 
Faced with this information vacuum, the reflection generator produces a 
plausible-sounding but causally wrong diagnosis. Unit tests provide step-level 
feedback --- which test case failed, which input produced the wrong output --- and 
this reduces confabulation from 32--82\% to 17\%. 
This suggests a systems-level mitigation that no prompt intervention can replicate: 
richer environment feedback. For ALFWorld specifically, this would mean returning 
explanations for ``Nothing happens'' responses rather than a bare failure signal.
 \subsection{Model Capability and Confabulation}

To test whether confabulation is an artifact of model 
weakness, we replicated standard Reflexion on all 16 frozen 
environments using gpt-4o-mini. Table~\ref{tab:gpt4o} summarizes 
the comparison.

\begin{table}[h]
\centering
\caption{gpt-3.5-turbo vs gpt-4o-mini on 16 frozen environments 
under standard Reflexion (no mitigation).}
\label{tab:gpt4o}
\begin{tabular}{lcc}
\toprule
                        & \textbf{gpt-3.5} & \textbf{gpt-4o-mini} \\
\midrule
Solved                  & 0/16             & 2/16             \\
Object mention rate     & 0\%              & 100\%            \\
Avg RRR                 & 0.64             & 0.53             \\
\bottomrule
\end{tabular}
\end{table}

gpt-4o-mini eliminates task-identity confabulation: 
all 142 reflections correctly name the target object, compared to 0/121 for gpt-3.5-turbo. Yet it solves only 2 of 16 frozen 
environments, identical to the no-memory ablation baseline. 
This confirms that task-identity confabulation and task-level 
capability are independent failure axes. A stronger model 
fixes the write-path failure (correct object in every 
reflection) but cannot fix the execution-step gap (the 13 
\texttt{pick\_*} environments remain unsolvable regardless of 
memory quality).

Notably, gpt-4o-mini introduces two new confabulation types absent 
in gpt-3.5-turbo. First, \emph{memory-format confabulation}: 
gpt-4o-mini generates structured numbered-list reflections 
(``1.~Clarify Task Requirements\ldots{} 2.~Verify Command 
Syntax\ldots'') that leak into action generation. The agent 
outputs planning text as actions, receiving \texttt{Nothing 
happens} for every step (env\_118, 9 consecutive trials). 
Second, when tested with gpt-4o-mini under enriched feedback, 
\emph{action-space confabulation} emerges: the agent generates 
natural-language actions (\texttt{check shelf 1 for book}) 
instead of valid ALFWorld syntax (\texttt{go to shelf 1}), 
looping on invalid commands for 47 steps (env\_97). These 
findings suggest that stronger models trade one confabulation 
type for another rather than eliminating the structural 
vulnerability.
\subsection{Generalization to Rule-Library Agents}
 
Memory confabulation is not a Reflexion-specific bug. It is a structural 
vulnerability of any agent that writes natural language to persistent memory based 
on self-assessment of binary failure signals. The common precondition is: 
\emph{binary feedback + self-generated reflection + persistent retrieval}.
 
ExpeL~\citep{zhao2024expel} faces this vulnerability with amplified consequences. 
Where Reflexion confabulates per-task reflections, ExpeL's rule extraction mechanism 
 which calls the same unconstrained LLM critique on failure trajectories  can produce globally-applied confabulated rules. A confabulated rule with two AGREE 
votes becomes entrenched and is applied across every evaluation environment, 
multiplying the harm of a single confabulation event.

\subsection{Limitations}

Our analysis has several limitations. We study confabulation 
exclusively in Reflexion, a single reflective agent architecture. 
While the structural vulnerability (binary feedback + 
self-generated reflection + persistent retrieval) should apply 
to any agent satisfying these conditions, empirical validation 
on other reflective systems such as ExpeL~\citep{zhao2024expel} 
or LATS \cite{zhou2024languageagenttreesearch} remains future work. Our gpt-4o-mini replication confirms 
the vulnerability persists across model sizes, but 
generalizability to other architectures cannot be assumed. The 
causal ablation is limited to 2 memory-harmful environments out 
of 16 frozen; while the effect is large (+6 and +7 trials), the 
small sample limits statistical power. Similarly, the HumanEval 
analysis covers only 4 frozen problems, making the cross-domain 
parallel structurally consistent but underpowered compared to the ALFWorld analysis (16 environments, 121 reflections). Our 
operational definition uses target-object mention as the 
grounding signal for detecting confabulation, which is a 
sufficient but not necessary condition: a reflection could 
mention the correct object while still misdiagnosing the failure 
cause, or confabulate in dimensions we do not measure such as 
wrong location or wrong action sequence. Finally, all experiments 
use pre-existing logs from the public Reflexion repository, 
fixing the model backbone (gpt-3.5-turbo), environment version, 
and hyperparameters; results may differ under other 
configurations.
\section{Conclusion}
 
We have shown that Reflexion agents systematically confabulate task identity in 
their reflective memory. Across 16 frozen environments and 121 reflections, the 
correct target object was mentioned zero times. Two environments require 7-8 trials with standard memory but solve in 1 trial without memory, providing causal evidence that stored memory can be actively harmful. Two grounding interventions prevented confabulation on solvable 
tasks but could not resolve hard-task failures, pointing to feedback granularity 
rather than reflection format as the fundamental bottleneck. The broader message for the agent memory community is that write-path validation is 
as important as retrieval quality. A memory system that stores confident, 
plausible-sounding but wrong beliefs is worse than no memory at all for the tasks 
those beliefs affect. Designing agents that know when not to write or that 
validate causal accuracy before storing is a necessary complement to the retrieval improvements that have dominated recent work.

\bibliography{example_paper}

@article{yao2022react,
  title={React: Synergizing reasoning and acting in language models},
  author={Yao, Shunyu and Zhao, Jeffrey and Yu, Dian and Du, Nan and Shafran, Izhak and Narasimhan, Karthik and Cao, Yuan},
  journal={arXiv preprint arXiv:2210.03629},
  year={2022}
}

@article{zhang2025survey,
  title={A survey on the memory mechanism of large language model-based agents},
  author={Zhang, Zeyu and Dai, Quanyu and Bo, Xiaohe and Ma, Chen and Li, Rui and Chen, Xu and Zhu, Jieming and Dong, Zhenhua and Wen, Ji-Rong},
  journal={ACM Transactions on Information Systems},
  volume={43},
  number={6},
  pages={1--47},
  year={2025},
  publisher={ACM New York, NY}
}

@misc{musser2008fastgenericsequencematching,
      title={A Fast Generic Sequence Matching Algorithm}, 
      author={David R. Musser and Gor V. Nishanov},
      year={2008},
      eprint={0810.0264},
      archivePrefix={arXiv},
      primaryClass={cs.DS},
      url={https://arxiv.org/abs/0810.0264}, 
}

@article{brown2020language,
  title={Language models are few-shot learners},
  author={Brown, Tom and Mann, Benjamin and Ryder, Nick and Subbiah, Melanie and Kaplan, Jared D. and Dhariwal, Prafulla and Neelakantan, Arvind and Shyam, Pranav and Sastry, Girish and Askell, Amanda and others},
  journal={Advances in Neural Information Processing Systems},
  volume={33},
  pages={1877--1901},
  year={2020}
}

@misc{zhou2024languageagenttreesearch,
      title={Language Agent Tree Search Unifies Reasoning Acting and Planning in Language Models}, 
      author={Andy Zhou and Kai Yan and Michal Shlapentokh-Rothman and Haohan Wang and Yu-Xiong Wang},
      year={2024},
      eprint={2310.04406},
      archivePrefix={arXiv},
      primaryClass={cs.AI},
      url={https://arxiv.org/abs/2310.04406}, 
}

@article{zhang2026feedback,
  title={Feedback-Driven Execution for LLM-Based Binary Analysis},
  author={Zhang, XiangRui and Li, Qiang and Wang, Haining},
  journal={arXiv preprint arXiv:2604.15136},
  year={2026}
}

@article{ji2023survey,
  title={Survey of hallucination in natural language generation},
  author={Ji, Ziwei and Lee, Nayeon and Frieske, Rita and Yu, Tiezheng and Su, Dan and Xu, Yan and Ishii, Etsuko and Bang, Ye Jin and Madotto, Andrea and Fung, Pascale},
  journal={ACM computing surveys},
  volume={55},
  number={12},
  pages={1--38},
  year={2023},
  publisher={ACM New York, NY}
}

@article{johnson1998false,
  title={False memories and confabulation},
  author={Johnson, Marcia K and Raye, Carol L},
  journal={Trends in cognitive sciences},
  volume={2},
  number={4},
  pages={137--145},
  year={1998},
  publisher={Elsevier}
}

@misc{maynez2020faithfulnessfactualityabstractivesummarization,
      title={On Faithfulness and Factuality in Abstractive Summarization}, 
      author={Joshua Maynez and Shashi Narayan and Bernd Bohnet and Ryan McDonald},
      year={2020},
      eprint={2005.00661},
      archivePrefix={arXiv},
      primaryClass={cs.CL},
      url={https://arxiv.org/abs/2005.00661}, 
}

@inproceedings{durante2025interactive,
  title={An interactive agent foundation model},
  author={Durante, Zane and Gong, Ran and Sarkar, Bidipta and Wake, Naoki and Taori, Rohan and Tang, Paul and Lakshmikanth, Shrinidhi and Schulman, Kevin and Milstein, Arnold and Vo, Hoi and others},
  booktitle={Proceedings of the Computer Vision and Pattern Recognition Conference},
  pages={3652--3662},
  year={2025}
}

@article{Ji_2023,
   title={Survey of Hallucination in Natural Language Generation},
   volume={55},
   ISSN={1557-7341},
   url={http://dx.doi.org/10.1145/3571730},
   DOI={10.1145/3571730},
   number={12},
   journal={ACM Computing Surveys},
   publisher={Association for Computing Machinery (ACM)},
   author={Ji, Ziwei and Lee, Nayeon and Frieske, Rita and Yu, Tiezheng and Su, Dan and Xu, Yan and Ishii, Etsuko and Bang, Ye Jin and Madotto, Andrea and Fung, Pascale},
   year={2023},
   month=Mar, pages={1–38} }

@misc{du2026memoryautonomousllmagentsmechanisms,
      title={Memory for Autonomous LLM Agents:Mechanisms, Evaluation, and Emerging Frontiers}, 
      author={Pengfei Du},
      year={2026},
      eprint={2603.07670},
      archivePrefix={arXiv},
      primaryClass={cs.AI},
      url={https://arxiv.org/abs/2603.07670}, 
}

@inproceedings{shinn2023reflexion,
  title     = {Reflexion: Language Agents with Verbal Reinforcement Learning},
  author    = {Shinn, Noah and Cassano, Federico and Berman, Edward and Gopinath, Ashwin and Narasimhan, Karthik and Yao, Shunyu},
  booktitle = {Advances in Neural Information Processing Systems},
  year      = {2023},
  url       = {https://openreview.net/forum?id=vAElhFcKW6}
}

@inproceedings{zhao2024expel,
  title     = {ExpeL: LLM Agents Are Experiential Learners},
  author    = {Zhao, Andrew and Huang, Daniel and Xu, Quentin and Lin, Matthieu and Liu, Yong-Jin and Huang, Gao},
  booktitle = {Proceedings of the AAAI Conference on Artificial Intelligence},
  volume    = {38},
  number    = {17},
  pages     = {19632--19642},
  year      = {2024},
  doi       = {10.1609/aaai.v38i17.29936},
  url       = {https://ojs.aaai.org/index.php/AAAI/article/view/29936}
}

@inproceedings{maynez2020faithfulness,
  title={On Faithfulness and Factuality in Abstractive Summarization},
  author={Maynez, Joshua and Narayan, Shashi and Bohnet, Bernd and McDonald, Ryan},
  booktitle={Proceedings of the 58th Annual Meeting of the Association for Computational Linguistics},
  pages={1906--1919},
  year={2020}
}

@article{chen2021evaluating,
  title={Evaluating Large Language Models Trained on Code},
  author={Chen, Mark and Tworek, Jerry and Jun, Heewoo and Yuan, Qiming and Pinto, Henrique Ponde de Oliveira and Kaplan, Jared and Edwards, Harri and Burda, Yuri and Joseph, Nicholas and Brockman, Greg and others},
  journal={arXiv preprint arXiv:2107.03374},
  year={2021}
}

@article{schnider2001spontaneous,
  title={Spontaneous confabulation, reality monitoring, and the limbic system—a review},
  author={Schnider, Armin},
  journal={Brain Research Reviews},
  volume={36},
  number={2-3},
  pages={150--160},
  year={2001},
  publisher={Elsevier}
}

@article{chrobak2009cognitive,
  title={The cognitive consequences of forced fabrication: Evidence from studies of eyewitness suggestibility},
  author={Chrobak, Q and Zaragoza, Maria S},
  journal={Confabulation: Views from neuroscience, psychiatry, psychology and philosophy},
  pages={67--90},
  year={2009},
  publisher={MIT Press Cambridge, MA}
}

@inproceedings{yang2018hotpotqa,
  title={HotpotQA: A dataset for diverse, explainable multi-hop question answering},
  author={Yang, Zhilin and Qi, Peng and Zhang, Saizheng and Bengio, Yoshua and Cohen, William and Salakhutdinov, Ruslan and Manning, Christopher D},
  booktitle={Proceedings of the 2018 conference on empirical methods in natural language processing},
  pages={2369--2380},
  year={2018}
}

@article{hurst2024gpt,
  title={Gpt-4o system card},
  author={Hurst, Aaron and Lerer, Adam and Goucher, Adam P and Perelman, Adam and Ramesh, Aditya and Clark, Aidan and Ostrow, AJ and Welihinda, Akila and Hayes, Alan and Radford, Alec and others},
  journal={arXiv preprint arXiv:2410.21276},
  year={2024}
}
\bibliographystyle{icml2026}

\newpage
\appendix
\onecolumn

\section{Prompt for Programmatic Extraction}
\label{sec:prompt}

The following prompt replaces the standard Reflexion self-diagnosis 
prompt. Failure steps are extracted programmatically from the 
trajectory before this prompt is constructed.

\begin{figure}[h]
\centering
\fbox{\parbox{0.9\columnwidth}{\small\ttfamily
You will be given the history of a past experience in which you 
were placed in an environment and given a task to complete. You 
were unsuccessful.\\[6pt]
YOUR TASK WAS: \{task\_line\}\\[6pt]
SPECIFIC FAILURES EXTRACTED FROM YOUR TRACE:\\
\hspace*{1em}Failure 1 (\{type\}):\\
\hspace*{2em}Action you took : \{action\}\\
\hspace*{2em}Environment said: \{response\}\\
\hspace*{1em}Failure 2 (\{type\}):\\
\hspace*{2em}Action you took : \{action\}\\
\hspace*{2em}Environment said: \{response\}\\[6pt]
Using the specific failures listed above (do not invent your 
own interpretation), explain in one sentence WHY each failure 
occurred. Then write a concise step-by-step New plan that 
avoids those exact failures. Name the exact objects and 
locations from your trace.\\[6pt]
Experience: \{scenario\}\\[6pt]
Previous plans:\\
\hspace*{1em}Attempt 1: \{plan\_1\}\\
\hspace*{1em}Attempt 2: \{plan\_2\}\\[6pt]
New plan:
}}
\caption{Programmatic extraction prompt template. Placeholders 
are filled from trajectory parsing before the LLM generates 
the reflection.}
\label{fig:prompt}
\end{figure}

The \texttt{\{task\_line\}} is extracted directly from the 
trajectory (the line containing ``Your task is to:''). The 
failure block is populated by a string parser that identifies 
actions receiving \texttt{Nothing happens} responses or 
repeated identical actions indicating a loop. The 
\texttt{\{scenario\}} is the full trajectory text. Previous 
plans are truncated to the action-relevant portion only, 
filtering out analysis text that could leak into action 
generation. This prompt structure ensures the agent receives 
concrete failure evidence rather than being asked to 
self-diagnose from an unstructured trajectory.


\end{document}